# Multilevel Browsing of Folksonomy-based Digital Collections

Joaquín Gayoso-Cabada, Daniel Rodríguez-Cerezo, José-Luis Sierra

Fac. Informática. Universidad Complutense de Madrid
C/ Prof. José García Santesmases 9,   28040 Madrid (Spain)
+34913947548

{jgayoso, drcerezo, jlsierra}@fdi.ucm.es

**Abstract.** This paper describes how to extend the usual one-level tag selection navigation paradigm in folksonomy-based digital collections to a *multilevel browsing* one, according to which it is possible to incrementally narrow down the set of selected objects in a collection by sequentially adding more and more filtering tags. For this purpose, we present a browsing strategy based on finite automata. Also, we provide some experimental results concerning the application of the approach in *Clavy*, a system for managing digital collections with reconfigurable structures in digital humanities and educational settings.

## 1. Introduction

Folksonomies are cataloguing schemes defined and applied collaboratively by communities of users. In this way, users not only apply folksonomies to organize digital resources, but they also actively contribute to their creation and maintenance [12]. In this context, to accommodate any but simplest interaction models can become a substantial technical challenge.

An example of interaction model particularly difficult to achieve is to implement general, unconstrained, *multi-level browsing* [5]. In this setting, a set of tags is sequentially picked up, and, in each stage, the set of objects tagged by all the selected tags is filtered. Even for collections of moderate size, to compute these sets of objects can be, in some cases, too costly to be achieved within acceptable response times. While by establishing predefined orders in which tags can be selected and by using these orders to create and maintain navigation trees the response times can be dramatically enhanced, this rigid and aprioristic organization is contrary to the dynamic and agile nature of folksonomies, where tag sets are continuously changing. In this paper we address this interaction style in its most unconstrained and general form.

The rest of the paper is organized as follows. Section 2 introduces the basis of folksonomy-like organizations of digital collections. Section 3 introduces the multilevel browsing paradigm for this kind of collections and describes how to enable such a browsing style efficiently. Section 4 presents some related work. Finally, section 5 outlines the final conclusions and some lines of future work.

| Tag cloud | 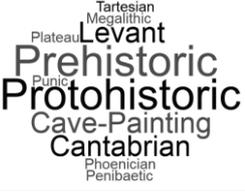 |
|---|---|
| Annotation | <table><tr><td>**Resource 1**</td><td>Cave-Painting, Cantabrian, Prehistoric</td><td>**Resource 4**</td><td>Tartesian, Plateau, Protohistoric</td></tr><tr><td>**Resource 2**</td><td>Cave-Painting, Levant, Prehistoric</td><td>**Resource 5**</td><td>Phoenician, Penibaetic, Protohistoric</td></tr><tr><td>**Resource 3**</td><td>Megalithic, Cantabrian, Prehistoric</td><td>**Resource 6**</td><td>Punic, Levant, Protohistoric</td></tr></table> |
| Resources | <table><tr><td>**Resource 1**</td><td>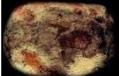</td><td>**Resource 4**</td><td>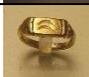</td></tr><tr><td>**Resource 2**</td><td>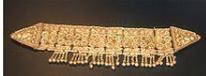</td><td>**Resource 5**</td><td>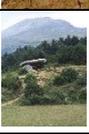</td></tr><tr><td>**Resource 3**</td><td>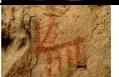</td><td>**Resource 6**</td><td>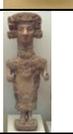</td></tr></table> |

Fig. 1. A small digital collection

## 2. Folksonomy-based Digital Collections

Collections organized with folksonomies typically comprise the following parts (see Fig. 1 for an example):
- On one hand, there are the *resources* of the collection. For instance, the small collection depicted in Fig. 1 includes six image archives as resources, corresponding to photographs of artistic objects from the Prehistoric and Protohistoric artistic periods in Spain (Fig. 1 actually shows thumbnails of these images).
- On the other hand, there is the *annotation* of the resources. This annotation consists of associating descriptive *tags* with resources. These tags are useful when cataloguing resources and, therefore, they enable future uses of the collection (navigation, search, etc.). For instance, in Fig. 1 the resource number 1 has the tags *Cave Painting*, *Cantabrian* and *Prehistoric* associated.
- Finally, there is a *tag cloud* that groups all the tags that can be used to annotate the resources. Thus, the tag cloud shown in Fig. 1 groups all the tags annotating resources in the collection. As usual, the size of tags in this cloud represents the presence (number of tagged resources) of the tag in the collection.

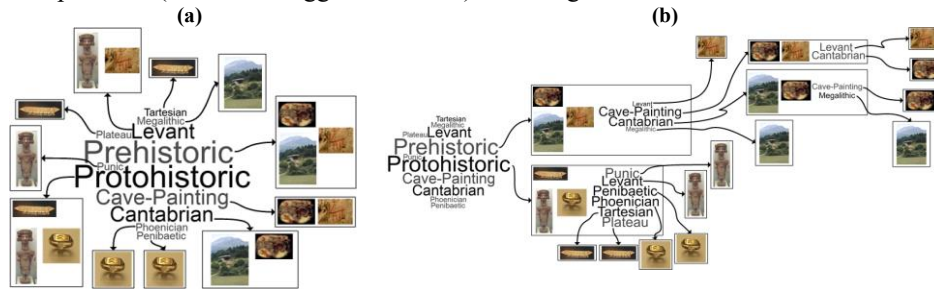

Fig. 2. Examples of (a) one-level browsing; (b) multi-level browsing

Thus, the internal organization of this kind of collection is very similar in appearance to classic keyword-based systems [15]. However, what distinguishes these collections from classic keyword-based systems is the social and inductive nature in the creation of the cataloguing schemata (i.e, the tag clouds). Indeed, folksonomy-based systems actively involve user communities that add, modify, delete and tag resources, using existing tags or creating new ones as needed. In this way, tag clouds are not explicitly defined nor explicitly maintained, but these emerge from the collaborative behavior of communities of practice [12]. While this somewhat uncontrolled and anarchic approach to tagging digital resources can additionally bring some relevant concerns and critics from a cataloguing point of view (e.g., existence of synonymous, irrelevant or very generic tags, etc.) [14], the fact is that these systems are extensively used in many scenarios (and especially in computer-mediated social ones) [3]. Thus, in this paper we will not focus on the critics and potential shortcomings of the approach, but on efficient ways of enabling sophisticated interaction strategies (multi-level browsing, in particular).

Folksonomy-like systems supports a simple one-level browsing strategy in a straightforward way. According to this strategy, it is possible to select one tag in the tag cloud and recover all the resources tagged with such a tag. Fig. 2(a) sketches this approach with the small collection of Fig. 1.

One-level browsing can be accomplished efficiently in a straightforward way by using and maintaining an *inverted index* [19], i.e., by associating with each tag references to the set of resources tagged with this tag, which directly bridges to the results of each selection. This simple and efficient implementation explains why most folksonomy-based systems include this interaction style as a primary browsing strategy. However it prevents more sophisticated exploratory behaviors involving simultaneously two or more tags. In the rest of the paper we will examine how to extend it to deal with more than a level.

## 3. Multilevel Browsing in Folksonomy-based Systems

This section addresses the multi-level browsing style in folksonomy-based systems. Subsection 3.1 introduces the basic interaction behavior. Subsection 3.2 characterizes this behavior as a finite state machine. Finally, subsection 3.3 gives some experimental results.

### 3.1. The Browsing Model

Conceptually, the extension of one-level to multi-level browsing in folksonomy-like systems is simple. Basically, when a tag is selected, not only the set of resources but also the tag cloud is narrowed down: the resulting tag cloud will be the *induced* by the set of filtered resources **R**. Such a tag will contain all the tags annotating some resource in **R** with the exception of those tags annotating *all* the resources in **R** (since, in this case, the selection would not refine the set of resources).It makes it possible to successively carry out new selections on the narrowed tag clouds until reaching a state containing an empty tag cloud. The expected behavior is partially illustrated in Fig. 2(b), which shows the set of resources and the associated tag clouds after some browsing actions on the collection in Fig. 1.

As in the case of one-level browsing, the multi-level browsing behavior can also be accomplished by using inverted indexes. However, now it is needed to evaluate conjunctive queries in each interaction state in order to determine the filtered resources. Although there was an extensive research on how to speed-up these operations [2], in some cases the time inverted can negatively impact on the interactive user experience.

### 3.2. Navigation automata

In order to speed-up multi-level browsing, it is necessary to have a suitable index structure. Ideally this structure should bridge to the set of resources selected by each meaningful set of tags $t_1$, ..., $t_n$, in the same way an inverted index directly provides the set of resources selected by a tag in the one-level approach. A way of providing such a structure is by using a finite state machine characterizing all the possible interactions and interaction states. This state machine will be called a *navigation automaton*. This automaton will consist of *states* labelled by set of resources, and *transitions* labelled by tags (as an example, Fig. 3a shows the navigation automaton for the collection of Fig. 1). More precisely:
- There will be an initial state labelled by all the resources in the collection.
- Given a state **S** labelled by a set of resources **R**, for each tag **t** in the tag cloud induced by **R** there will be a state **S'** labelled by all the resources in **R** annotated by **t**, as well as a transition from **S** to **S'** labelled by **t**.

Since the navigation automaton embeds all the possible ways of multi-level navigation, it can support multi-level browsing in a straightforward way. Unfortunately, in some cases the number of states in this automaton can grow very fast (in the worst case, exponentially with respect to the number of resources). The most extreme case, in which the number of states is $2^n-1$ (with *n* the number of resources), arises, for instance, by distinguishing each pair of resource annotations in a single tag. In order to avoid this potential exponential factor of the explicit construction of navigation automata, it is possible to maintain non-deterministic versions of these automata, in such a way only states representing a disjoint partition of their parent states are maintained. Fig. 3b shows a feasible non-deterministic automaton equivalent to the shown in Fig. 3a (it is worthwhile to notice this solution may not be unique).

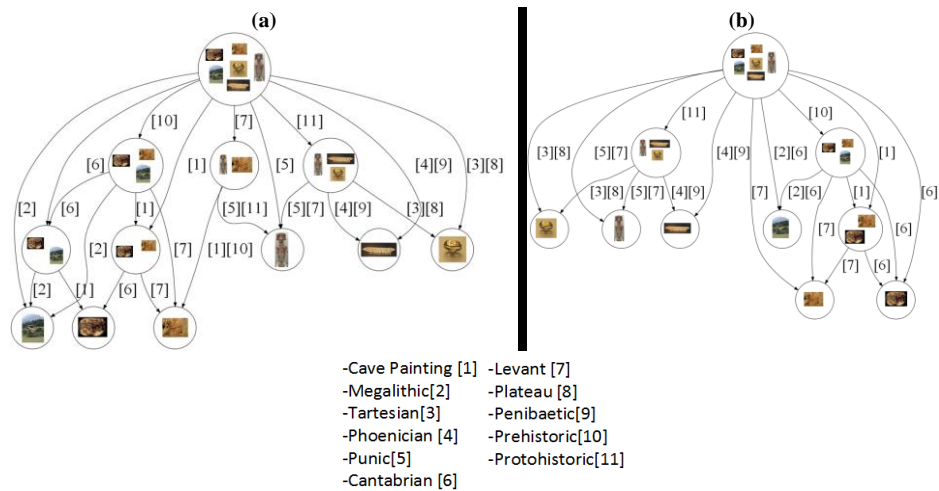

-Cave Painting [1]  -Levant [7]
-Megalithic[2]      -Plateau [8]
-Tartesian[3]       -Penibaetic[9]
-Phoenician [4]     -Prehistoric[10]
-Punic[5]           -Protohistoric[11]
-Cantabrian [6]

Fig. 3. (a) Navigation automaton for the collection in Fig. 1; (b) A non-deterministic version of the automaton in (a)

### 3.3. Experimental Evaluation

In order to evaluate our multilevel browsing approach, we have implement it on *Clavy*, an experimental system for managing digital collections that lets users define organization schemata in a collaborative way. [1] In order to provide some structure for facilitating navigation, *Clavy* makes it possible to group tags in categories organized hierarchically. Nevertheless, this hierarchy is not pre-established, but it can be edited by *Clavy* users at any time (see Fig. 4). Therefore, in the backstage, multi-level browsing support in *Clavy* must be resorted to the basic model described in section 3, since the hierarchy is also object of continuous change and evolution. In addition to the automata-based browsing framework described in this paper, we have also implemented an inverted index-based solution in *Clavy*, using Lucene [13], a robust and highly optimized framework for implementing information retrieval applications.

In this context, we set up an experiment consisting of adding the resources in *Chasqui* [17],[2] a digital collection of 6283 digital resources on Precolombian

---
[1] http://clavy.fdi.ucm.es/Clavy/
[2] http://oda-fec.org/ucm-chasqui

American archeology, to *Clavy* and to simulate runs concerning hierarchy reconfiguration and browsing operations.

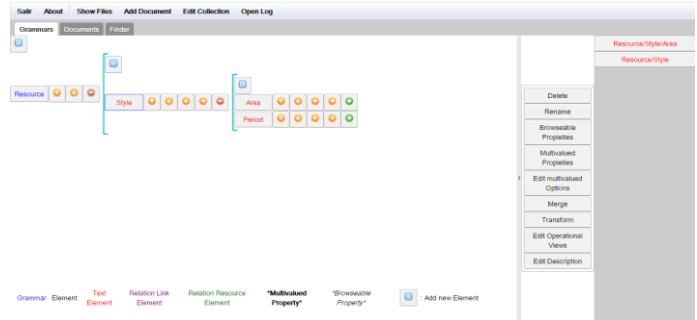

Fig. 4. Editing a hierarchy of tag categories with *Clavy*.

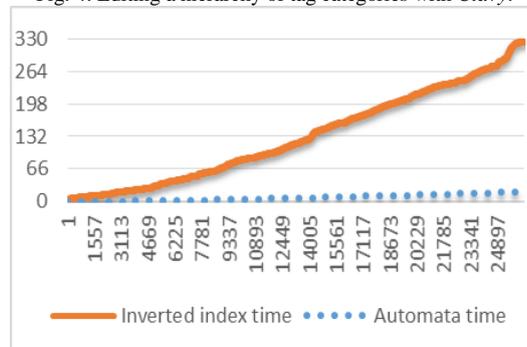

Fig. 5. Cumulative time of inverted indexes vs. automata

Each run was customized as follows. We interleaved resource insertion with hierarchy reconfiguration / browsing rounds. Each insertion round consisted of 100 resource insertions (with the exception of the last one, in which the remaining resources where inserted). In turn, each browsing / reconfiguration round consisted of executing $0.1n$ browsing operations randomly interleaved with $0.01n$ reconfigurations ($n$ being the number of resources inserted so far). Each browsing operation consisted, in turn, of selecting a feasible tag and computing the next set of active objects, or in establishing the initial state as the active one in case of unavailability of feasible tags; once determined the next interaction state, all the filtered resources were visited. In both cases of inverted indexes and automata, in-memory indexes were used in order to avoid side effects of persistence, disturbing the experiment.

Fig. 5 shows the results obtained from the two runs. The experiment was run on a PC with Windows 10, with a 3.4GHz Intel microprocessor, and with 8Gb of DDR3 RAM. Horizontal axis corresponds to the number of operations carried out so far. Vertical axis corresponds to cumulative time (in seconds). As is makes apparent, automata-based approach clearly outperforms inverted indexes one (regardless we are using a highly optimized framework, like Lucene, for inverted indexing vs. our own in-house experimental implementation for navigation automata).

## 4. Related Work

There are several systems that, like our proposal, implement several sorts of multi-level browsing on folksonomy-based systems. Systems like the described in [7][9] are supported by inverted indexes approaches. Other systems, like the described in [11], are supported by extensible data adapters that interfaces between synchronized tag clouds and underlying database management systems. Instead of relying on inverted indexes and/or conventional database layers, our approach starts by characterizing the intrinsic behavior of multi-level browsing on folksonomy-like system in terms of navigation automata, and then proposes to approximate this model with a non-deterministic version that provides reasonable time and space tradeoffs. In [4] we propose a representation of these non-deterministic automata inspired in *dendrograms* as those used in hierarchical clustering settings [8].

Our navigation automata model is actually similar to lattice-based proposals to browse information spaces, as described in the seminal work of [5]. This organization is actually the main subject of the fertile theory of *formal concept analysis* [16]. Similarly, there are several proposals on using lattices as the underlying indexing structures for enabling multi-level browsing [6][18]. However, all these approaches are limited by the intrinsic complexity of formal concept analysis [10]. It is why we proposed a simpler but still practical approximation based on non-deterministic versions of navigation automata.

## 5. Conclusions and Future Work

Folksonomy-based digital collections are live entities in which not only digital resources, but also organization schemata, are object of continuous change and evolution. This changing and evolving nature makes the accomplishment of sophisticated interaction paradigms particularly challenging. In this paper we have addressed the efficient inclusion of multilevel browsing strategies in these settings, in which sets of selected resources can be successively refined through the selection of sequences of tags. For this purpose we have modeled this behavior as a finite state machine, the *navigation automaton*, taking into account all the possible ways of navigating the collection using tags. Unfortunately, we have also showed how, in some cases, the number of states of this automaton can increase exponentially with respect to the collection's size. In order to address this potential exponential factor we have proposed to use non-deterministic versions of these automata. Some experiments with a real collection gave us evidence on how the automata-based technique can outperform more conventional and widely used ones, like those based on inverted indexes.

We are currently working on further optimizing our navigation automata representation. We are also looking for efficient ways to persist all this information, either by using standard relational databases or alternative NoSQL approaches. Finally, we also hope to include support for arbitrary Boolean queries and for alternative ways of exploring the resources selected.

## Acknowledgements

This work has been supported by the BBVA Foundation (grant HUM14_251) and Spanish Ministry of Economy and Competitiveness (grant TIN2014-52010-R)